\documentclass[runningheads]{llncs}
\usepackage{graphicx}
\usepackage{courier}

\usepackage{tikz}
\usepackage{comment}
\usepackage{amsmath,amssymb} 
\usepackage{color}

\begin{document}
\pagestyle{headings}
\mainmatter
\def\ECCVSubNumber{98}  

\title{DeepGIN: Deep Generative Inpainting Network for Extreme Image Inpainting} 

\titlerunning{DeepGIN for Extreme Image Inpainting}
%
\author{Chu-Tak Li\inst{1} \and
Wan-Chi Siu\inst{1} \and
Zhi-Song Liu\inst{2}
\and
Li-Wen Wang\inst{1} \and
Daniel Pak-Kong Lun\inst{1}}
\authorrunning{Chu-Tak Li et al.}
%
\institute{Centre for Multimedia Signal Processing, Department of Electronic and Information Engineering, The Hong Kong Polytechnic University 
\and
LIX, Ecole Polytechnique, CNRS, IP Paris, France}
\maketitle

\begin{abstract}
The degree of difficulty in image inpainting depends on the types and sizes of the missing parts. Existing image inpainting approaches usually encounter difficulties in completing the missing parts in the wild with pleasing visual and contextual results as they are trained for either dealing with one specific type of missing patterns (mask) or unilaterally assuming the shapes and/or sizes of the masked areas. We propose a deep generative inpainting network, named DeepGIN, to handle various types of masked images. We design a Spatial Pyramid Dilation (SPD) ResNet block to enable the use of distant features for reconstruction. We also employ Multi-Scale Self-Attention (MSSA) mechanism and Back Projection (BP) technique to enhance our inpainting results. Our DeepGIN outperforms the state-of-the-art approaches generally, including two publicly available datasets (FFHQ and Oxford Buildings), both quantitatively and qualitatively. We also demonstrate that our model is capable of completing masked images in the wild.
\keywords{Image Inpainting, Attention, Back Projection}
\end{abstract}

\section{Introduction}

Image inpainting (also called image completion) is a task of predicting the values of missing pixels in a corrupted/masked image such that the completed image looks realistic and is semantically close to the reference ground truth even though it does not exist in real-world situations. This task would be useful for repairing corrupted photos or erasing unwanted parts from photos. It could also serve applications of restoration of photos and footage of films, scratch removal, automatic modifications to images and videos, and so forth. Because of the wide-ranging applications, image inpainting has been an overwhelming research topic in the computer vision and graphics communities for decades. 

\begin{figure}[t]
\centering
\includegraphics[width=0.95\textwidth]{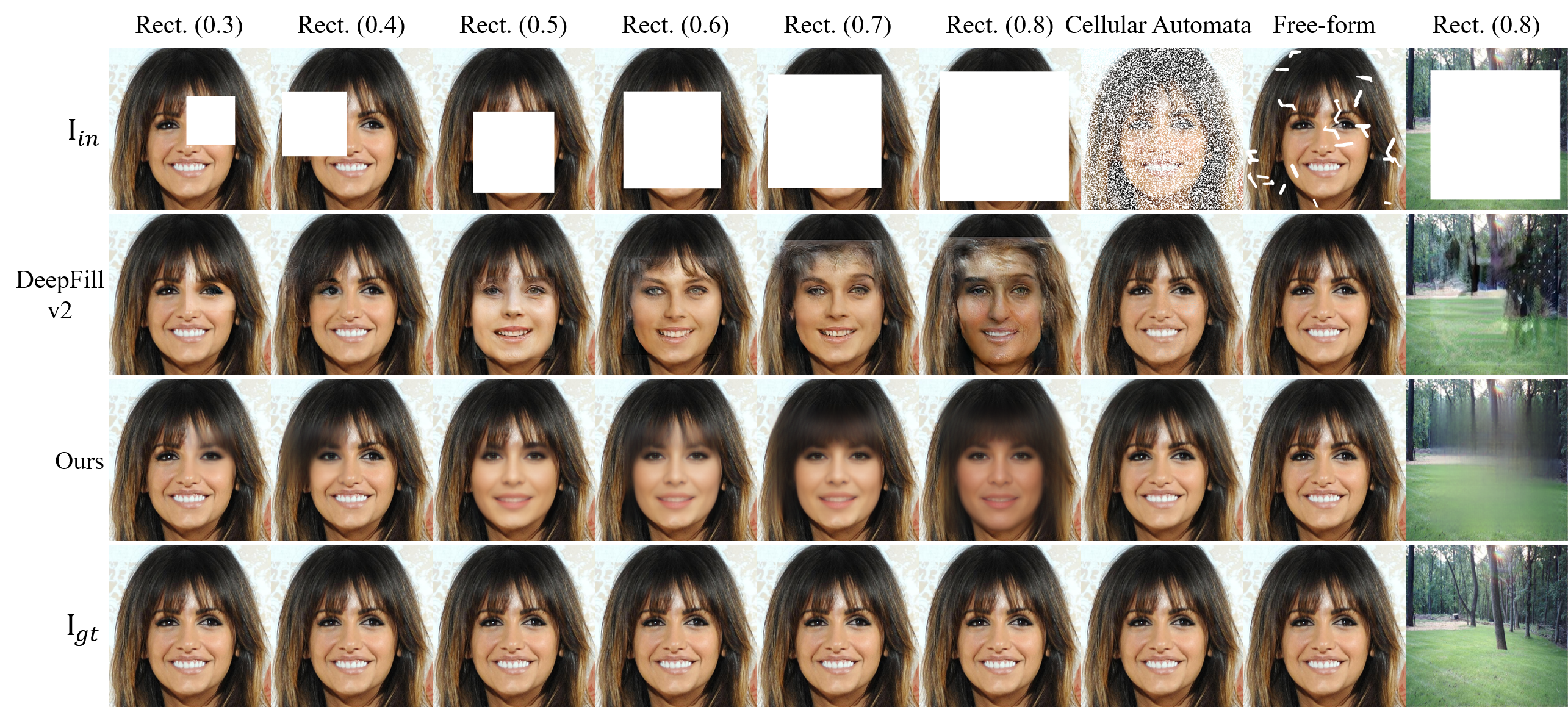}
\vspace{-1\baselineskip}
\caption{\textbf{Degree of difficulty in extreme image inpainting.} From top to bottom: the first row shows the input masked images $\textbf{I}_{in}$ with the corresponding mask described on top of them. Rect. ($\alpha$) represents a random rectangular mask with the height and width rate of $\alpha$ of each dimension. The randomly generated mask based on cellular automata is introduced in the AIM 2020 Extreme Image Inpainting Challenge \cite{AIM2020}, and the free-form mask is proposed in DeepFillv2 \cite{Deepfillv2}. The second and third rows are the completed images using DeepFillv2 and our proposed DeepGIN respectively. The last row displays the ground truth images. Please zoom in to see the examples at the $6^{\mathrm{th}}$ and the last column especially}
\label{fig:poi}
\vspace{-1.5\baselineskip}
\end{figure}

Inspired by the recent success of deep learning approaches at the tasks of image recognition \cite{NonLocalBlk,ResNet}, image super-resolution \cite{AIM19-SR,NTIRE20-SR,ESRGAN}, visual place recognition and localization \cite{ITSC,ToDayGAN}, image enlightening \cite{DLN}, image synthesis \cite{pix2pix,pix2pixHD} and many others, a growing number of CNN based methods of image inpainting \cite{CE,MSNPS,GLCIC,CA,PartialConv,EdgeConnect,Deepfillv2} have been proposed to fill images with holes in an end-to-end manner. For example, Iizuka et al. \cite{GLCIC} employed dilated convolutions instead of standard convolutions to widen the receptive field at each layer for better conservation of the spatial structure of an image. Yu et al. \cite{CA} proposed a two-stage generative network with a contextual attention layer to intentionally consider correlated feature patches at distant spatial locations for coherent estimation of local missing pixels. Liu et al. \cite{PartialConv} suggested a partial convolutional layer to identify the non-hole regions at each layer such that the convolutional results are derived only from the valid pixels. However, the effectiveness of these strategies depends highly on the scales and forms of the missing regions as well as the contents of both the valid and invalid pixels as shown in Fig.~\ref{fig:poi}. Based on this observation, we aim for a generalized inpainting network which can complete masked images in the wild. 

In this paper, we present a coarse-to-fine Deep Generative Inpainting Network (DeepGIN) which consists of two stages, namely coarse reconstruction stage and refinement stage. Similar to the network design of previous studies \cite{CA,EdgeConnect,Deepfillv2}, the coarse reconstruction stage is responsible for rough estimation of the missing pixels in an image while the refinement stage is responsible for detailed decoration on the coarse reconstructed image. In order to obtain a realistic and coherent completed image, Spatial Pyramid Dilation (SPD), Multi-Scale Self-Attention (MSSA) and Back Projection (BP) techniques are redesigned and embedded in our proposed network. The main function of SPD is to extensively allow for different receptive fields such that information gathered from both surrounding and distant spatial locations can contribute to the prediction of local missing regions. The concept of SPD is applied to both stages while MSSA and BP are integrated into the refinement stage. The core idea of MSSA is that it takes the self-similarity of the image itself at multiple levels into account for the coherence on the completed image while BP enforces the alignment of the predicted and given pixels in the completed image. 

The contributions made in this work are summarized as follows: 
\begin{itemize}
    \item We propose a Spatial Pyramid Dilation (SPD) block to deal with different types of masks with various shapes and sizes. 
    \item We stress the importance of self-similarity to image inpainting and we significantly improve our inpainting results by employing the strategy with Multi-Scale Self-Attention (MSSA).
    \item We design a Back Projection (BP) strategy for obtaining the inpainting results with better alignment of the generated patterns and the reference ground truth.
\end{itemize}

\vspace{-1\baselineskip}
\section{Related Work}
\vspace{-0.5\baselineskip}

Existing approaches to image inpainting can be classified into two categories, namely conventional and deep learning based methods. PatchMatch \cite{PatchMatch} is one of the representative conventional methods in which similar patches from a target image or other source images are copied and pasted into the missing regions of the target image. However, it is computationally expensive even a fast approximate nearest-neighbor search algorithm has been adopted to alleviate the problem of costly patch search iterations. 

\vspace{-1\baselineskip}
\subsubsection{Regular Mask.} Context Encoder \cite{CE} is the first deep learning based inpainting algorithm that employs the framework of Generative Adversarial Networks (GANs) \cite{GANs} for more realistic image completion. For GAN based image inpainting, a generator is designed for filling the missing regions with semantic awareness and a discriminator is responsible for distinguishing the completed image and the reference ground truth. Based on this setting, the generator and discriminator are alternately optimized to compete against each other, as a result, the completed image given by the generator would be visually and semantically close to the reference ground truth. Specifically, Pathak et al. \cite{CE} resize images to 128$\times$128 and assume a 64$\times$64 rectangular center missing region for the task of inpainting. The encoded feature of the image with the center hole is then decoded to reconstruct a 64$\times$64 image for the center hole.

Based on this early work, Yang et al. \cite{MSNPS} attached the pre-trained VGG-19 \cite{VGG-19} as their proposed texture network to perform the task in a coarse-to-fine manner. The output of the context encoders is then passed to the texture network for local texture refinement to enhance the accuracy of the reconstructed hole content. Iizuka et al. \cite{GLCIC} suggested an approach in which two auxiliary discriminators are designed for ensuring both the global and local consistency of the completed image. The global discriminator takes the entire image as input for differentiation between real and completed images while the local discriminator examines only the local area around the filled region. To further alleviate the visual artifacts in the filled region, they also employed Poisson image blending as a simple post-processing. Expanding on this idea and the above-mentioned idea of PatchMatch \cite{PatchMatch}, Yu et al. \cite{CA} also adopted a two-stage coarse-to-fine approach with global and local discriminators. A contextual attention layer is proposed and applied to the second refinement network which plays the similar role of the post-processing. 

\vspace{-1\baselineskip}
\subsubsection{Irregular Mask.} For the early stage of deep learning based methods of image inpainting, authors focused on the rectangular types of masks and this assumption limits the effectiveness of these methods in real-world situations. Liu et al. \cite{PartialConv} addressed this problem by suggesting a partial convolutional layer, in which a binary mask for indicating the missing regions is automatically updated along with the convolutional operations inside their model for guiding the reconstruction. Nazeri et al. \cite{EdgeConnect} forced an image completion network to generate images with fine-grained details by providing guidance for filling the missing regions with the use of their proposed edge generator. The edge generator is responsible for predicting a full edge map of the masked image. With the estimated edge map as additional information, their trained model can be extended to an interactive image editing tool in which users can sketch the outline of the missing regions to obtain tailor-made completed images. Combining the concept of partial convolution with optional user-guided image inpainting, Yu et al. \cite{Deepfillv2} improved their previous model \cite{CA} by proposing gated convolution for free-form image inpainting. They modify the hard-assigned binary mask in partial convolution to a learnable soft-gated convolutional layer. The soft gating layer can be achieved by using convolutional filters with size 1$\times$1 followed by a sigmoid function. However, additional soft gating layers introduce additional parameters and the effectiveness still depends on the scales of the masked areas.

Our work echoes the importance of information given by distant spatial locations and self-similarity of the image itself to image inpainting. We increase the number of receptive fields and apply multi-scale self-attention strategy to handle various types of masks in the wild. Our multi-scale self-attention strategy is derived from the non-local network \cite{NonLocalBlk}, in which the correlation between features is emphasized and it has been used in image super-resolution \cite{AIM19-SR,APBN}. For achieving better coherency of the completed images, we also adopt the back projection technique \cite{DBPN,APBN} to encourage better alignment of the generated and valid pixels. We weight the back projected residual instead of using the parametric back projection blocks as in \cite{DBPN,APBN} to avoid more additional parameters.

\vspace{-0.5\baselineskip}
\section{Problem Formulation}
\vspace{-0.5\baselineskip}

Let us start to define an input RGB masked image and a binary mask image as $\textbf{I}_{in} \in\mathbb{R}^{H\times W\times 3}$ and $\textbf{M} \in\mathbb{R}^{H\times W}$ respectively. The pixel values input to our model are normalized between 0 and 1 and pixels with value 1 in $\textbf{M}$ represent the masked regions. $\textbf{I}_{coarse} \in\mathbb{R}^{H\times W\times 3}$ denotes the output of our coarse generator $\textit{G}_{1}$ at the first coarse reconstruction stage. We also define the output of our refinement generator $\textit{G}_{2}$ at the second refinement stage and the reference ground truth image as $\textbf{I}_{out} \in\mathbb{R}^{H\times W\times 3}$ and $\textbf{I}_{gt} \in\mathbb{R}^{H\times W\times 3}$ respectively. Note that $\textit{H}$ and $\textit{W}$ are the height and width of an input/output image and we fix the input to 256$\times$256 for inpainting. Our objective is straightforward. We would like to complete $\textbf{I}_{in}$ conditioned on $\textbf{M}$ and produce a completed image $\textbf{I}_{out}$ ($\textbf{I}_{compltd}$) which should be both visually and semantically close to the reference ground truth $\textbf{I}_{gt}$. $\textbf{I}_{compltd}$ is the same as $\textbf{I}_{out}$ except the valid pixels are directly replaced by the ground truth. We propose a coarse-to-fine network trained under the framework of generative adversarial learning with training data $\{\textbf{I}_{in},\textbf{M},\textbf{I}_{gt}\}$ where $\textbf{M}$ is randomly generated with arbitrary sizes and shapes. Generator $\textit{G}_{1}$ takes $\textbf{I}_{in}$ and $\textbf{M}$ as input and generates $\textbf{I}_{coarse}$ as output. Subsequently, we feed $\textbf{I}_{coarse}$ and $\textbf{M}$ to generator $\textit{G}_{2}$ to obtain the completed image $\textbf{I}_{out}$ ($\textbf{I}_{compltd}$). 

\begin{figure}[t]
\centering
\includegraphics[width=\textwidth]{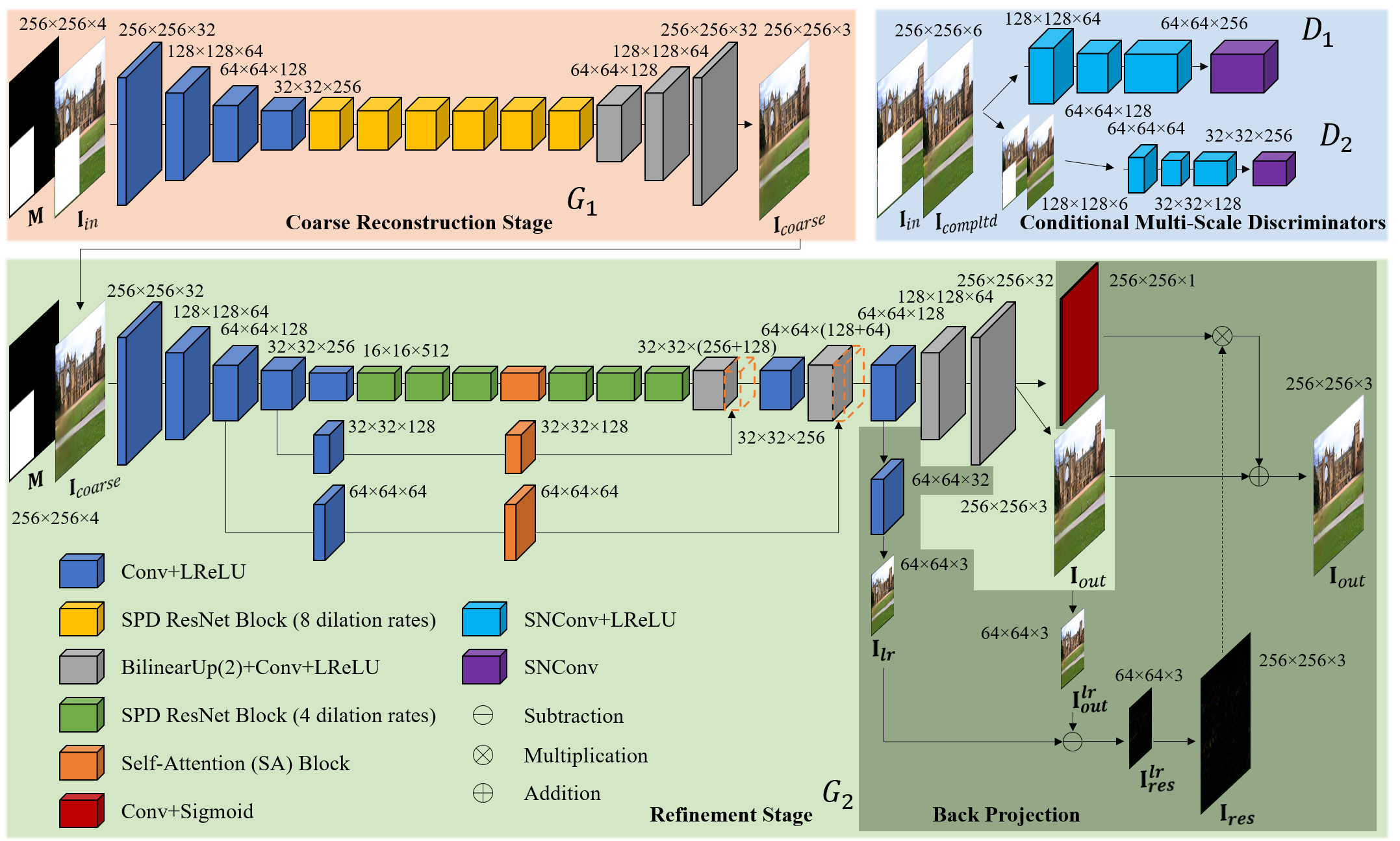}
\vspace{-1\baselineskip}
\caption{\textbf{Architecture of our proposed model for image inpainting.} Our proposed model consists of two generators and two discriminators. The coarse generator $\textit{G}_{1}$ at Coarse Reconstruction Stage and the second refinement generator $\textit{G}_{2}$ at Refinement Stage constitute our DeepGIN which is used in both training and testing. The two discriminators $\textit{D}_{1}$ and $\textit{D}_{2}$ located within Conditional Multi-Scale Discriminators area are only used in training as an auxiliary network for generative adversarial training}
\label{fig:architecture}
\vspace{-1.5\baselineskip}
\end{figure}

\vspace{-0.75\baselineskip}
\section{Approach}
\vspace{-0.75\baselineskip}

Our proposed Deep Generative Inpainting Network (DeepGIN) consists of two stages as shown in Fig.~\ref{fig:architecture}, a coarse reconstruction stage and a refinement stage. The first coarse generator $\textit{G}_{1}(\textbf{I}_{in},\textbf{M})$ is trained to roughly reconstruct the masked regions and gives $\textbf{I}_{coarse}$. The second refinement generator $\textit{G}_{2}(\textbf{I}_{coarse},\textbf{M})$ is trained to exquisitely decorate the coarse prediction with details and textures, and eventually forms the completed image $\textbf{I}_{out}$ ($\textbf{I}_{compltd}$). For our discriminators, motivated by SN-GANs \cite{SN-GANs,Deepfillv2} and multi-scale discriminators \cite{pix2pix,pix2pixHD}, we modify and employ two SN-GAN based discriminators $\textit{D}(\textbf{I}_{in},\textbf{I}_{compltd})$ which operate at two image scales, 256$\times$256 and 128$\times$128 respectively, to encourage better details and textures of local reconstructed patterns at different scales. Details of our network architecture and learning are shown below.

\vspace{-1\baselineskip}
\subsection{Network Architecture}

\subsubsection{Coarse Reconstruction Stage.}

\begin{figure}[t]
\centering
\includegraphics[width=\textwidth]{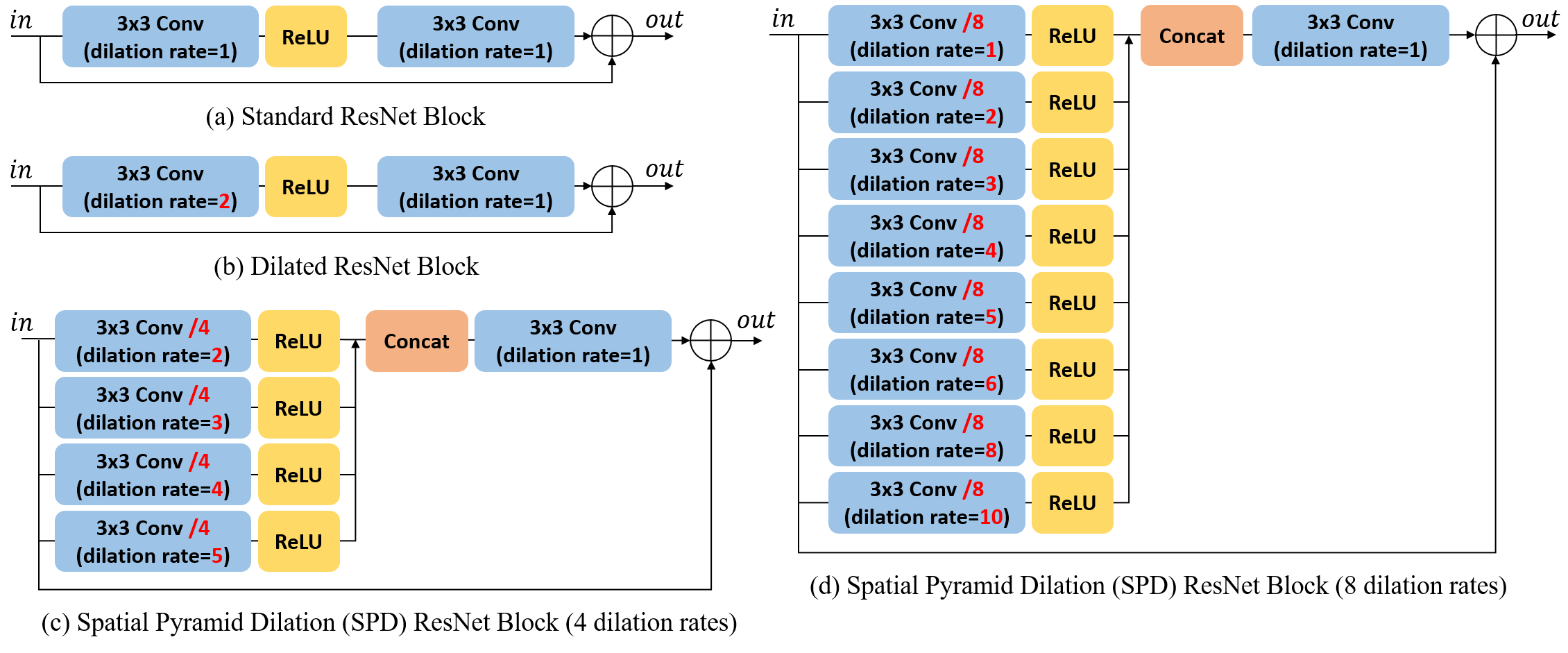}
\vspace{-1.5\baselineskip}
\caption{\textbf{Variations of ResNet Block.} From top to bottom, left to right: (a) Standard ResNet block \cite{ResNet}, (b) Dilated ResNet block used in \cite{GLCIC,CA,EdgeConnect} which adopts a dilation rate of 2 of the first convolutional layer, (c) The proposed SPD ResNet block with 4 dilation rates and (d) 8 dilation rates. To avoid additional parameters, we split the number of input feature channels into equal parts according to the number of dilation rates employed. As shown in (c), if 4 dilation rates are used, the output channel size of the first convolutions equals a quarter of the input channel size}
\label{fig:spd}
\vspace{-1.5\baselineskip}
\end{figure}

Recall that $\textit{G}_{1}$ is our coarse generator and it is responsible for rough estimation of the missing pixels in a masked image. Referring to the previous section, we concatenate $(\textbf{I}_{in},\textbf{M}) \in \mathbb{R}^{H\times W\times (3+1)}$ as the input to $\textit{G}_{1}$ and then obtain the coarse image $\textbf{I}_{coarse}$. $\textit{G}_{1}$ follows an encoder-decoder structure. As the scales of the masked regions are randomly determined, we propose a Spatial Pyramid Dilation (SPD) ResNet block with various dilation rates to enlarge the receptive fields such that information given by distant spatial locations can be included for reconstruction. Our SPD ResNet block is an improved version of the original ResNet block \cite{ResNet} as shown in Fig.~\ref{fig:spd}, and in total, 6 SPD ResNet blocks with 8 different dilation rates are used at this stage. 

\vspace{-1\baselineskip}
\subsubsection{Refinement Stage.}

Generator $\textit{G}_{2}$ is designed for refinement of $\textbf{I}_{coarse}$ and it is similar to generator $\textit{G}_{1}$. At this stage, we have 6 SPD ResNet blocks with 4 different dilation rates and a Self-Attention (SA) block in between at the middle layers. Apart from the SPD ResNet block, Multi-Scale Self-Attention (MSSA) blocks \cite{NonLocalBlk,APBN} are used for self-similarity consideration. The SA block used in this paper is exactly the same as the one proposed in \cite{NonLocalBlk}. One similarity between the SA block and the contextual attention layer \cite{CA,Deepfillv2} is that they both have the concept of self-similarity which is useful for amending the reconstructed patterns based on the remaining ground truth in a masked image. We apply MSSA instead of single scale SA to enhance the coherency of the completed image $\textbf{I}_{out}$ by attending on the self-similarity of the image itself at three different scales, namely 16$\times$16, 32$\times$32 and 64$\times$64 as shown in Fig.~\ref{fig:architecture}. To avoid an excessive increase in additional parameters, we simply use standard convolutional layers to reduce the channel size before connecting to the SA blocks. The idea of Back Projection (BP) \cite{DBPN,APBN} is also redesigned and it is used at the last decoding process of this stage (see the shaded Back Projection region in Fig.~\ref{fig:architecture}). At the layer with spatial size of 64$\times$64, we output a low-resolution (LR) completed image $\textbf{I}_{lr}$ and perform BP with $\textbf{I}_{out}$. By learning to weight the BP residual and adding it back to update $\textbf{I}_{out}$, the generated patterns can have better alignments with the reference ground truth and hence $\textbf{I}_{out}$ looks more coherent. 

\vspace{-1\baselineskip}
\subsubsection{Conditional Multi-Scale Discriminators.}

Two discriminators $\textit{D}_{1}$ and $\textit{D}_{2}$ at two input scales (i.e. 256$\times$256 and 128$\times$128) are trained together with the generators to stimulate details of the filled regions. Combining the idea of multi-scale discriminators \cite{pix2pixHD} with SN-GANs \cite{SN-GANs} and PatchGAN \cite{pix2pix,Deepfillv2}, our $\textit{D}_{1}(\textbf{I}_{in},\textbf{I})$ and $\textit{D}_{2}(\textbf{I}_{in},\textbf{I})$ take the concatenation result of two RGB images as input ($\textbf{I}$ is either $\textbf{I}_{compltd}$ or $\textbf{I}_{gt}$, recall that $\textbf{I}_{compltd}$ is the same as $\textbf{I}_{out}$ except the valid pixels are directly replaced by the ground truth) and output a set of feature maps with size of $H/2^2\times W/2^2\times c$ where $\textit{c}$ represents the number of feature maps. Note that each value on these output feature maps represents a local region in the input image at two different scales. By training $\textit{D}_{1}$ and $\textit{D}_{2}$ to discriminate between real and fake local regions, $\textbf{I}_{out}$ would gradually be close to its reference ground truth $\textbf{I}_{gt}$ in terms of both appearance and semantic similarity. For achieving stable generative adversarial learning, we employ the spectral normalization layer described in \cite{SN-GANs} after each convolutional layer in $\textit{D}_{1}$ and $\textit{D}_{2}$. 

\vspace{-1\baselineskip}
\subsection{Network Learning}
\vspace{-0.5\baselineskip}

We design our loss function based on consideration to both quantitative accuracy and visual quality of the completed images. Our loss function consists of five major terms, namely (i) a $\textit{L1 loss}$ to ensure the pixel-wise reconstruction accuracy especially if using quantitative evaluation metrics such as PSNR and mean L1 error to evaluate the completed images; (ii) an $\textit{adversarial loss}$ to urge the distribution of the completed images to be close to the distribution of the real images; (iii) the $\textit{feature perceptual loss}$ used in \cite{Johnson} that encourages each completed image and its reference ground truth image to have similar feature representations as computed by a well-trained network with good generalization like VGG-19 \cite{VGG-19}; (iv) the $\textit{style loss}$ \cite{StyleTransfer} to emphasize the style similarity such as textures and colors between completed images and real images; and (v) the $\textit{total variation loss}$ used as a regularizer in \cite{Johnson} to guarantee the smoothness in the completed images by penalizing its visual artifacts or discontinuities.

\vspace{-1\baselineskip}
\subsubsection{L1 Loss.} 

Our $\textit{L1 loss}$ is derived from three image pairs, namely $\textbf{I}_{coarse}$ and $\textbf{I}_{gt}$; $\textbf{I}_{out}$ and $\textbf{I}_{gt}$; and $\textbf{I}_{lr}$ and $\textbf{I}^{lr}_{gt}$. Note that $\textbf{I}^{lr}_{gt}$ is obtained by down-sampling $\textbf{I}_{gt}$ by 4 times. We sum the L1-norm distances of these three image pairs and define our $\textit{L1 loss}$, $\mathcal{L}_{L1}$, as follows:
\vspace{-0.75\baselineskip}
\begin{align}
  \mathcal{L}_{L1} = \lambda_{hole} \mathcal{L}_{hole} + \mathcal{L}_{valid}
\vspace{-0.75\baselineskip}
\end{align}
where $\mathcal{L}_{hole}$ and $\mathcal{L}_{valid}$ are the sums of the distances which are calculated only from the missing pixels and the valid pixels respectively. $\lambda_{hole}$ is a weight to the pixel-wise loss within the missing regions.

\vspace{-1\baselineskip}
\subsubsection{Adversarial Loss.}

For generative adversarial learning, our discriminators are trained to rightly distinguish $\textbf{I}_{compltd}$ from $\textbf{I}_{gt}$ while our generators strive to cheat the discriminators of incorrect classification. We employ the hinge loss to train our model, $\mathcal{L}_{adv,G}$ and $\mathcal{L}_{adv,D}$ are computed as:
\vspace{-0.5\baselineskip}
\begin{align}
\vspace{-1\baselineskip}
  \mathcal{L}_{adv,G} = - \mathbb{E}_{\textbf{I}_{in} \sim \mathbb{P}_{i}} [\textit{D}_{1}(\textbf{I}_{in},\textbf{I}_{compltd})] - \mathbb{E}_{\textbf{I}_{in} \sim \mathbb{P}_{i}} [\textit{D}_{2}(\textbf{I}_{in},\textbf{I}_{compltd})] 
\end{align}
\vspace{-1.5\baselineskip}
\begin{align}
  \mathcal{L}_{adv,D} = \mathbb{E}_{\textbf{I}_{in} \sim \mathbb{P}_{i}} \left[ \sum_{d=1}^{2} [\text{ReLU}(1-\textit{D}_{d}(\textbf{I}_{in},\textbf{I}_{gt})) +
  \text{ReLU}(1+\textit{D}_{d}(\textbf{I}_{in}, \textbf{I}_{compltd}))] \right] 
\vspace{-1\baselineskip}
\end{align}
where $\mathbb{P}_{i}$ represents the data distribution of $\textbf{I}_{in}$, ReLU is the rectified linear unit defined as $f(x) = \max(0,x)$.

\vspace{-1\baselineskip}
\subsubsection{Perceptual Loss.}

Let $\phi$ be the well-trained loss network, VGG-19 \cite{VGG-19}, and $\phi_{l}^{\textbf{I}}$ be the activation maps of the $\textit{l}^{\mathrm{th}}$ layer of the network $\phi$ given an image $\textbf{I}$. We choose five layers of the pre-trained VGG-19, namely $\textit{conv}1\_1$, $\textit{conv}2\_1$, $\textit{conv}3\_1$, $\textit{conv}4\_1$, and $\textit{conv}5\_1$ for computing this loss. Our $\mathcal{L}_{perceptual}$ is calculated as:
\vspace{-0.5\baselineskip} 
\begin{align}
\vspace{-1\baselineskip}
  \mathcal{L}_{perceptual} = \sum_{l=1}^{L} \frac{\| \phi_{l}^{\textbf{I}_{out}} - \phi_{l}^{\textbf{I}_{gt}} \|_{1}}{N_{\phi^{\textbf{I}_{gt}}_{l}}} + \sum_{l=1}^{L} \frac{\| \phi_{l}^{\textbf{I}_{compltd}} - \phi_{l}^{\textbf{I}_{gt}} \|_{1}}{N_{\phi^{\textbf{I}_{gt}}_{l}}} 
\vspace{-1\baselineskip}  
\end{align}
where $N_{\phi^{\textbf{I}_{gt}}_{l}}$ indicates the number of elements in $\phi^{\textbf{I}_{gt}}_{l}$ and $\textit{L}$ equals 5 as five layers are used. Here, we compute the L1-norm distance between the high-level feature representations of $\textbf{I}_{out}$, $\textbf{I}_{compltd}$ and $\textbf{I}_{gt}$ given by the network $\phi$.

\vspace{-1\baselineskip}
\subsubsection{Style Loss.}

Let $(\phi_{l}^{\textbf{I}})^\top (\phi_{l}^{\textbf{I}})$ be the Gram matrix \cite{StyleTransfer} which computes the feature correlations between each activation map of the $\textit{l}^{\mathrm{th}}$ layer of $\phi$ given $\textbf{I}$, and this is also called auto-correlation matrix. We then calculate the $\textit{style loss}$ ($\mathcal{L}_{style}$) using $\textbf{I}_{out}$, $\textbf{I}_{compltd}$ and $\textbf{I}_{gt}$ as:
\vspace{-0.5\baselineskip}
\begin{align}
\vspace{-0.5\baselineskip}
  \mathcal{L}_{style} = \sum_{\textbf{I}}^{\textbf{I}_{out},\textbf{I}_{compltd}} \sum_{l=1}^{L} \frac{1}{C_{l}C_{l}} \left|\left| \frac{1}{C_{l}H_{l}W_{l}} ((\phi_{l}^{\textbf{I}})^\top (\phi_{l}^{\textbf{I}}) - (\phi_{l}^{\textbf{I}_{gt}})^\top (\phi_{l}^{\textbf{I}_{gt}})) \right|\right|_{1}
\vspace{-1\baselineskip}
\end{align}
where $C_{l}$ denotes the number of activation maps of the $\textit{l}^{\mathrm{th}}$ layer of $\phi$. $H_{l}$ and $W_{l}$ are the height and width of each activation map of the $\textit{l}^{\mathrm{th}}$ layer of $\phi$. Note that we use the same five layers of the VGG-19 as mentioned for this loss as well. 

\vspace{-1\baselineskip}
\subsubsection{Total Variation (TV) Loss.}

We also adopt the total variation regularization to ensure the smoothness in $\textbf{I}_{compltd}$.
\vspace{-0.5\baselineskip}
\begin{align}
\vspace{-1\baselineskip}
  \mathcal{L}_{tv} = \sum_{x,y}^{H-1,W} \frac{\| \textbf{I}_{compltd}^{x+1,y} - \textbf{I}_{compltd}^{x,y} \|_{1}}{N^{row}_{\textbf{I}_{compltd}}}  + \sum_{x,y}^{H,W-1} \frac{\| \textbf{I}_{compltd}^{x,y+1} - \textbf{I}_{compltd}^{x,y} \|_{1}}{N^{col}_{\textbf{I}_{compltd}}} 
\vspace{-1.5\baselineskip}
\end{align}
where $\textit{H}$ and $\textit{W}$ are the height and width of $\textbf{I}_{compltd}$. $\textit{N}^{row}_{\textbf{I}_{compltd}}$ and $\textit{N}^{col}_{\textbf{I}_{compltd}}$ are the number of pixels in $\textbf{I}_{compltd}$ except for the last row and the last column respectively. 

\vspace{-1\baselineskip}
\subsubsection{Total Loss.} Our total loss function for the generators is the weighted sum of the five major loss terms:
\vspace{-0.5\baselineskip}
\begin{align}
\vspace{-1\baselineskip}
  \mathcal{L}_{total} = \mathcal{L}_{L1} + \lambda_{adv} \mathcal{L}_{adv,G} + \lambda_{perceptual} \mathcal{L}_{perceptual} + \lambda_{style} \mathcal{L}_{style} + \lambda_{tv} \mathcal{L}_{tv}
\vspace{-1\baselineskip}
\end{align}
where $\lambda_{adv}$, $\lambda_{perceptual}$, $\lambda_{style}$, and $\lambda_{tv}$ are the hyper-parameters which indicate the significance of each term.

\vspace{-0.75\baselineskip}
\section{Experimental Work}
\vspace{-0.75\baselineskip}

We have participated in the AIM 2020 Extreme Image Inpainting Challenge \cite{AIM2020} of the ECCV 2020 (please find in our github page for details and qualitative results of the challenge). In designing our proposed model, we take reference to the networks in \cite{Johnson,pix2pix,pix2pixHD}. We have attached our improved SPD ResNet block to our DeepGIN. We have also modified and applied the ideas of MSSA and BP in our proposed model. Inspired by ESRGAN \cite{ESRGAN}, we remove all batch normalization layers in the model to smooth out the related visual artifacts. We have used discriminators at two different scales which share the same architecture. Also, we have adjusted the number of layers of each discriminator and applied spectral normalization layers \cite{SN-GANs,Deepfillv2} after the convolutional layers for training stability.
\vspace{-1\baselineskip}

\subsection{Training Procedure}
\vspace{-0.25\baselineskip}

\subsubsection{Random Mask Generation.} Three different types of masks are used in our training. The first type is a rectangular mask with the height and width between 30-70$\%$ of each dimension \cite{AIM2020,CE,MSNPS,GLCIC,CA}. The second type is the free-form mask proposed in \cite{Deepfillv2}. The third type of masks is introduced in the AIM 2020 Image Inpainting Challenge \cite{AIM2020}, for which masks are randomly generated based on cellular automata. During training, each mask was randomly generated and we applied the three types of masks to each training image to get three different masked images. We observed that this can balance the three types of masks to achieve more stable training. 

\vspace{-1\baselineskip}
\subsubsection{Training Batch Formation.} As the size of training images could be very diverse, we resized all training images to the size of 512$\times$512 and adopted a sub-sampling method \cite{PixelShuffle} to randomly select a sub-image with size of 256$\times$256. We then apply the random mask generation as stated above to obtain three masked images. Therefore, each training image becomes three training images. We set a batch size of 4 and this means that there are 12 training images in a batch.

\vspace{-1\baselineskip}
\subsubsection{Two-Stage Training.} Our training process is divided into two stages, namely a warm-up stage and then the main stage. First, we trained only the generators by using the $\textit{L1 loss}$ for 10 epochs. We used the initialization method mentioned in \cite{ESRGAN}, using a smaller initialization for ease of training a very deep network. The trained model at the warm-up stage was used as an initialization for the main stage. This $\textit{L1}$-oriented pre-trained model provides a reasonable initial point for training GANs, for which a balance between quantitative accuracy of the reconstruction and visual quality of the output is required. For the main stage, we trained the generators alternately with the discriminators for 100 epochs. We used Adam \cite{Adam} with momentum 0.5 for both stages. The initial learning rates for generators and discriminators were set to 0.0001 and 0.0004 respectively. We trained them for 10 epochs with the initial rates and linearly decayed the rates to zero over the last 90 epochs. The hyper-parameters of the loss terms in Eq. (1) and Eq. (7) were set to $\lambda_{hole}$ = 5.0, $\lambda_{adv}$ = 0.001, $\lambda_{perceptual}$ = 0.05, $\lambda_{style}$ = 80.0, and $\lambda_{tv}$ = 0.1. We developed our model using Pytorch 1.5.0 \cite{Pytorch} and trained it on two NVIDIA GeForce RTX 2080Ti GPUs.

\vspace{-1\baselineskip}
\subsection{Training Data}
\vspace{-0.25\baselineskip}

\subsubsection{ADE20K Dataset.} We trained our model on the subset of ADE20K dataset \cite{ADE20K1,ADE20k2} for participating in the AIM challenge \cite{AIM2020}. This dataset is collected for scene parsing and understanding, in which it contains images from various scene categories. The subset is provided by the organizers of the challenge and it consists of 10,330 training images with diverse resolutions roughly, from 256$\times$256 to 3648$\times$2736. We took around two and a half days for training on this dataset.

\vspace{-1\baselineskip}
\subsubsection{CelebA-HQ Dataset.} Beyond the ADE20K dataset, we also trained our model on the CelebA-HQ dataset \cite{CelebA-HQ} that contains 30K high-quality face images with a standard size of 1024$\times$1024. We randomly split this dataset into two groups, 27,000 images for training and 3,000 images for testing. This required approximately 6 days to train our model on this dataset.

\vspace{-0.75\baselineskip}
\section{Analysis of Experimental Results}
\vspace{-0.75\baselineskip}

We have thoroughly evaluated our proposed model. We first provide evidence in our model analysis to show the effectiveness of our suggested strategies for using Spatial Pyramid Dilation (SPD) ResNet block, Multi-Scale Self-Attention (MSSA), and Back Projection (BP). We then compare our model with state-of-the-art approaches, namely DeepFillv1 \cite{CA} and DeepFillv2 \cite{Deepfillv2}, which are known to have a good generalization. We demonstrate that our model is able to handle images in the wild by testing it on two publicly available datasets, namely Flickr-Faces-HQ (FFHQ) dataset \cite{FFHQ} and The Oxford Buildings (Oxford) dataset \cite{Oxford}. Related materials are available at: \texttt{https://github.com/rlct1/DeepGIN}.

\vspace{-1\baselineskip}
\subsection{Model Analysis}
\vspace{-0.25\baselineskip}

We first evaluate the effectiveness of the three proposed strategies, namely SPD, MSSA, and BP. Refer to the proposed architecture as shown in Fig.~\ref{fig:architecture}, our baselines are denoted as StdResBlk (Coarse only, using only the Coarse Reconstruction Stage) and StdResBlk (a conventional ResNet for inpainting), for which all SA blocks and BP branch are eliminated and all SPD ResNet blocks are replaced by standard ResNet blocks (see Fig.~\ref{fig:spd}(a)). DilatedResBlk or SPDResBlk represents StdResBlk with standard ResNet blocks replaced by Dilated or SPD ResNet blocks. Please refer to Fig.~\ref{fig:spd}(b),(c), and (d). SA or MSSA indicates whether single SA block or MSSA is used and the use of BP is denoted as BP. We conducted the model analysis on CelebA-HQ dataset \cite{CelebA-HQ} using the 3K testing images. Note that the testing images were randomly masked by the three types of masks and the same set of masked images was used for each variation of our model. During testing, for images with size larger than 256$\times$256, we divided the input into a number of 256$\times$256 sub-images using the sub-sampling method \cite{PixelShuffle} and obtained the completed sub-images. We then regrouped the sub-images to form the completed image by using the reverse sub-sampling method. We finally replaced the valid pixels by the ground truth.

\setlength{\tabcolsep}{14pt}
\begin{table}[t]
\begin{center}
\caption{Model analysis of our proposed model on CelebA-HQ dataset. The best results are in \textbf{bold} typeface}
\vspace{-0.5\baselineskip}
\label{table:ablation}
\scalebox{0.7}
{
\begin{tabular}{lcccccl}
\hline
\multicolumn{1}{c}{\begin{tabular}[c]{@{}c@{}}Variations of\\ our model\end{tabular}} & \begin{tabular}[c]{@{}c@{}}Number of\\ parameters\end{tabular} & PSNR           & SSIM           & L1 err.(\%)   & FID             & LPIPS \\ \hline \hline
StdResBlk (Coarse only)                                                                & 8.168M                                                         & 31.55          & 0.925          & 4.690          & 23.824          & 0.182      \\
StdResBlk                                                                              & 40.850M                                                        & 31.34          & 0.923          & 4.710          & 19.436          & 0.191      \\
StdResBlk-SA                                                                           & 41.376M                                                        & 31.60          & 0.925          & 4.510          & 18.239          & 0.180      \\
StdResBlk-MSSA                                                                         & 42.892M                                                        & 32.66          & 0.933          & 4.067          & 12.843          & 0.148      \\
DilatedResBlk-MSSA                                                                     & 42.892M                                                        & 32.71          & 0.933          & 4.034          & 12.548          & 0.149      \\
SPDResBlk-MSSA                                                                         & 42.892M                                                        & 32.88          & 0.935          & 3.884          & 12.335          & 0.143      \\
SPDResBlk-MSSA-BP                                                                      & 42.930M                                                        & \textbf{33.26} & \textbf{0.939} & \textbf{3.666} & \textbf{11.424} & \textbf{0.132}      \\ \hline \hline
\end{tabular}
}
\end{center}
\end{table}
\vspace{-1\baselineskip}
\setlength{\tabcolsep}{1.4pt}

\begin{figure}[t]
\centering
\vspace{-1.5\baselineskip}
\includegraphics[width=0.95\textwidth]{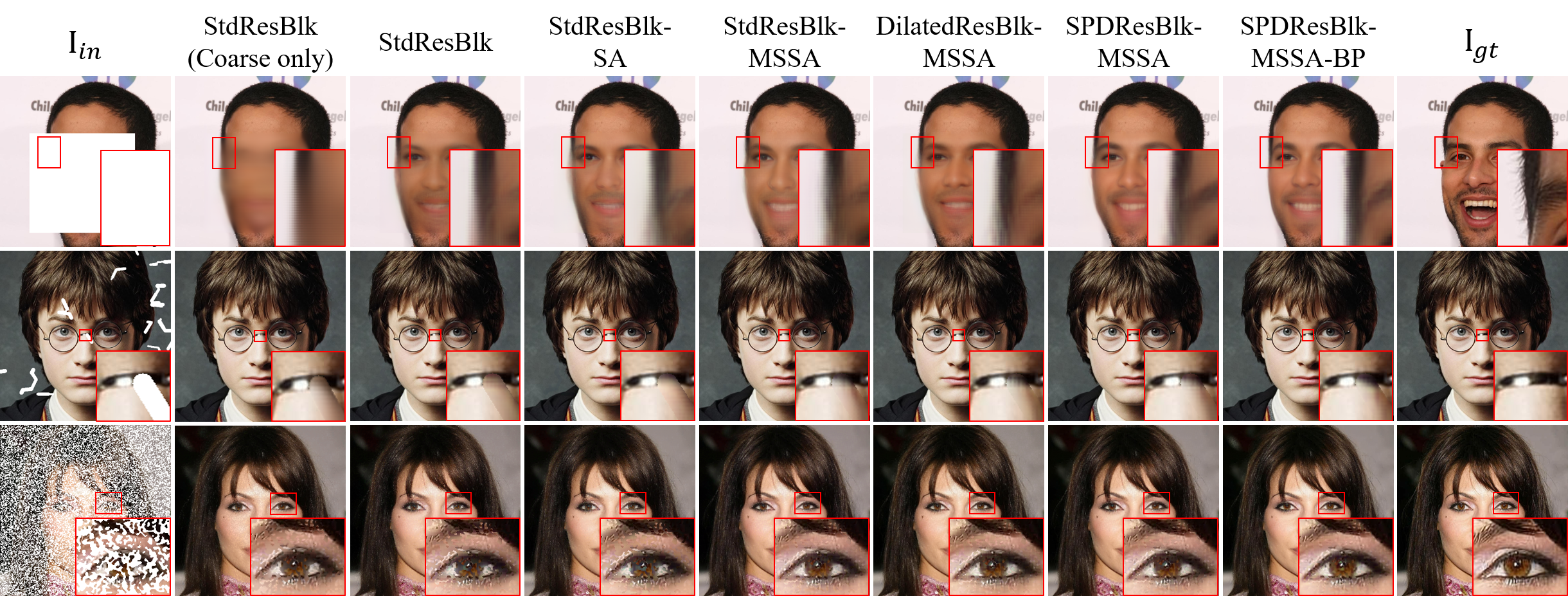}
\vspace{-1\baselineskip}
\caption{\textbf{Comparisons of test results of the variations of our model on CelebA-HQ dataset.} Three different types of masked images are displayed from top to bottom. The first and the last columns show $\textbf{I}_{in}$ and $\textbf{I}_{gt}$ respectively. The variations of our model are indicated on top of Fig.\ref{fig:ablation_study}. Our full model (the $8^{\mathrm{th}}$ column), SPDResBlk-MSSA-BP (GAN based), provides high quality results with both the best similarity and visual quality to the ground truth images. Please zoom in for a better view}
\label{fig:ablation_study}
\vspace{-1.5\baselineskip}
\end{figure}

\vspace{-0.25\baselineskip}
\subsubsection{Quantitative Comparisons.} As the lack of good quantitative evaluation metric for inpainting \cite{CA,PartialConv,Deepfillv2}, we report several numerical metrics which are commonly used in image manipulation, namely PSNR, SSIM \cite{SSIM}, mean L1 error, Fr\'{e}chet Inception Distance (FID) \cite{FID}, and Learned Perceptual Image Patch Similarity (LPIPS) \cite{LPIPS}, for a comprehensive analysis of the performance. The results are listed in Table~\ref{table:ablation} and higher PSNR, SSIM and smaller L1 err. mean better pixel-wise reconstruction accuracy. FID and LPIPS are also used to estimate the visual quality of the output, the smaller the better. It is obvious that our full model, SPDResBlk-MSSA-BP, gives the best performance on these numerical metrics. The employment of MSSA brings an 1.06 dB increase in PSNR compared to StdResBlk-SA. This reflects the importance of multi-scale self-similarity to inpainting. Our SPD ResNet blocks and the adoption of BP also bring about 0.22 dB and 0.38 dB improvement in PSNR respectively.

\vspace{-1.25\baselineskip}
\subsubsection{Qualitative Comparisons.} Fig.~\ref{fig:ablation_study} shows the comparisons of the variations of our model on CelebA-HQ dataset. Without the second refinement stage, the completed images lack for facial details like the first example of the $2^{\mathrm{nd}}$ column in Fig.~\ref{fig:ablation_study}. It can also be observed that the use of MSSA greatly enhances the visual quality as compared to the two which are without SA block and with only a single SA block (see the $3^{\mathrm{rd}}$ and $4^{\mathrm{th}}$ columns). Apart from this, with the SPD ResNet blocks and BP technique, the completed images are with better color coherency and alignment of the generated features. For example, see the spectacle frames in the $2^{\mathrm{nd}}$ row.

\setlength{\tabcolsep}{22pt}
\begin{table}[t]
\begin{center}
\caption{Comparisons of DeepFillv1 \cite{CA} and DeepFillv2 \cite{Deepfillv2} on both FFHQ and Oxford datasets with two sets of masked images. One set only contains the rectangular masks while another set includes all the three types of masks. Our DeepGIN is denoted as Ours (i.e. the full model, SPDResBlk-MSSA-BP in the previous section). (OS) and (256) mean that the testing images are with the original sizes and size of 256$\times$256 respectively. The best results are in \textbf{bold} typeface}
\vspace{-0.8\baselineskip}
\label{table:comparisons}
\scalebox{0.7}
{
\begin{tabular}{lccccc}
\hline
\multicolumn{1}{c}{Method}     & PSNR     & SSIM     & L1 err.(\%)  & FID     & LPIPS  \\ \hline \hline
\multicolumn{6}{c}{Flickr-Faces-HQ Dataset (FFHQ), random rectangular masks}           \\ \hline \hline
DeepFillv1 (OS)                & 20.22    & 0.872    & 16.523       & 97.630  & 0.173  \\
DeepFillv2 (OS)                & 20.95    & 0.903    & 14.607       & 92.070  & 0.170  \\
Ours (OS)                      & \textbf{26.05}    & \textbf{0.923}    & \textbf{7.183}        & \textbf{20.849}  & \textbf{0.137}  \\ \hline
DeepFillv1 (256)               & 21.55    & 0.836    & 13.631       & 26.276  & 0.144  \\
DeepFillv2 (256)               & 22.52    & 0.845    & 12.029       & \textbf{19.336}  & \textbf{0.128}  \\
Ours (256)                     & \textbf{24.36}    & \textbf{0.867}    & \textbf{9.797}        & 37.577  & 0.142  \\ \hline \hline
\multicolumn{6}{c}{The Oxford Buildings Dataset (Oxford), random rectangular masks}    \\ \hline \hline
DeepFillv1 (OS)                & 19.20    & 0.767    & 20.322       & 67.193  & 0.187  \\
DeepFillv2 (OS)                & 18.58    & 0.766    & 21.204       & 77.636  & 0.192  \\
Ours (OS)                      & \textbf{21.92}    & \textbf{0.861}    & \textbf{12.067}       & \textbf{63.744}  & \textbf{0.170}  \\ \hline
DeepFillv1 (256)               & 19.49    & 0.795    & 16.851       & \textbf{58.588}  & \textbf{0.169}  \\
DeepFillv2 (256)               & 18.88    & 0.789    & 18.308       & 66.615  & 0.174  \\
Ours (256)                     & \textbf{21.90}    & \textbf{0.819}    & \textbf{12.995}       & 74.866  & 0.185  \\ \hline \hline
\multicolumn{6}{c}{Flickr-Faces-HQ Dataset (FFHQ), random three types of masks}        \\ \hline \hline
DeepFillv1 (OS)                & 25.12    & 0.839    & 11.363       & 64.534  & 0.232  \\
DeepFillv2 (OS)                & 29.70    & 0.912    & 7.994        & 36.940  & 0.188  \\
Ours (OS)                      & \textbf{32.36}    & \textbf{0.929}    & \textbf{4.071}        & \textbf{14.327}  & \textbf{0.156}  \\ \hline
DeepFillv1 (256)               & 22.87    & 0.683    & 16.812       & 80.952  & 0.310  \\
DeepFillv2 (256)               & 22.75    & 0.716    & 17.472       & 75.555  & 0.293  \\
Ours (256)                     & \textbf{24.71}    & \textbf{0.760}    & \textbf{13.417}       & \textbf{64.542}  & \textbf{0.274}  \\ \hline \hline
\multicolumn{6}{c}{The Oxford Buildings Dataset (Oxford), random three types of masks} \\ \hline \hline
DeepFillv1 (OS)                & 21.48    & 0.741    & 23.460       & 61.958  & 0.237  \\
DeepFillv2 (OS)                & 24.68    & 0.802    & 19.195       & 38.315  & \textbf{0.179}  \\
Ours (OS)                      & \textbf{27.57}    & \textbf{0.871}    & \textbf{7.268}        & \textbf{38.016}  & 0.191  \\ \hline
DeepFillv1 (256)               & 21.64    & 0.686    & 18.835       & 81.009  & 0.284  \\
DeepFillv2 (256)               & 20.80    & 0.702    & 20.687       & 82.671  & 0.266  \\
Ours (256)                     & \textbf{23.60}    & \textbf{0.744}    & \textbf{14.659}       & \textbf{79.927}  & \textbf{0.265}  \\ \hline \hline
\end{tabular}
}
\end{center}
\vspace{-2.5\baselineskip}
\end{table}
\setlength{\tabcolsep}{1.4pt}

\vspace{-1\baselineskip}
\subsection{Comparison With Previous Works}
\vspace{-0.5\baselineskip}

In order to test the generalization of our model, we compare our best model against some state-of-the-art approaches, DeepFillv1 \cite{CA} and DeepFillv2 \cite{Deepfillv2}, on the two publicly available datasets, FFHQ \cite{FFHQ} and Oxford Buildings \cite{Oxford}. It is worth noting that both DeepFillv1 and v2 are known to have good generalization for dealing with images in the wild. We directly used their provided pre-trained models\footnote{\texttt{https://github.com/JiahuiYu/generative\_inpainting}} for comparison. The FFHQ dataset is similar to the CelebA-HQ dataset and it contains 70K high-quality face images at 1024$\times$1024 resolution. We randomly selected 1,000 images for the testing on this dataset. For the Oxford dataset, it consists of 5,062 images of Oxford landmarks with a wide variety of styles. The images include buildings, suburban areas, halls, people, etc. We also randomly selected 523 testing images on this dataset for comparison.

Similarly, testing images were randomly masked by the three types of masks. For DeepFillv1 and v2, the authors divided an image into a number of grids to perform inpainting and indicated that their models were trained with images of resolution 256$\times$256. Note also that DeepFillv1 was trained only for the rectangular types of masks. For fair comparison, we also conducted experiments in which testing images were randomly masked only by the rectangular masks.

\begin{figure}[t]
\centering
\includegraphics[width=0.88\textwidth]{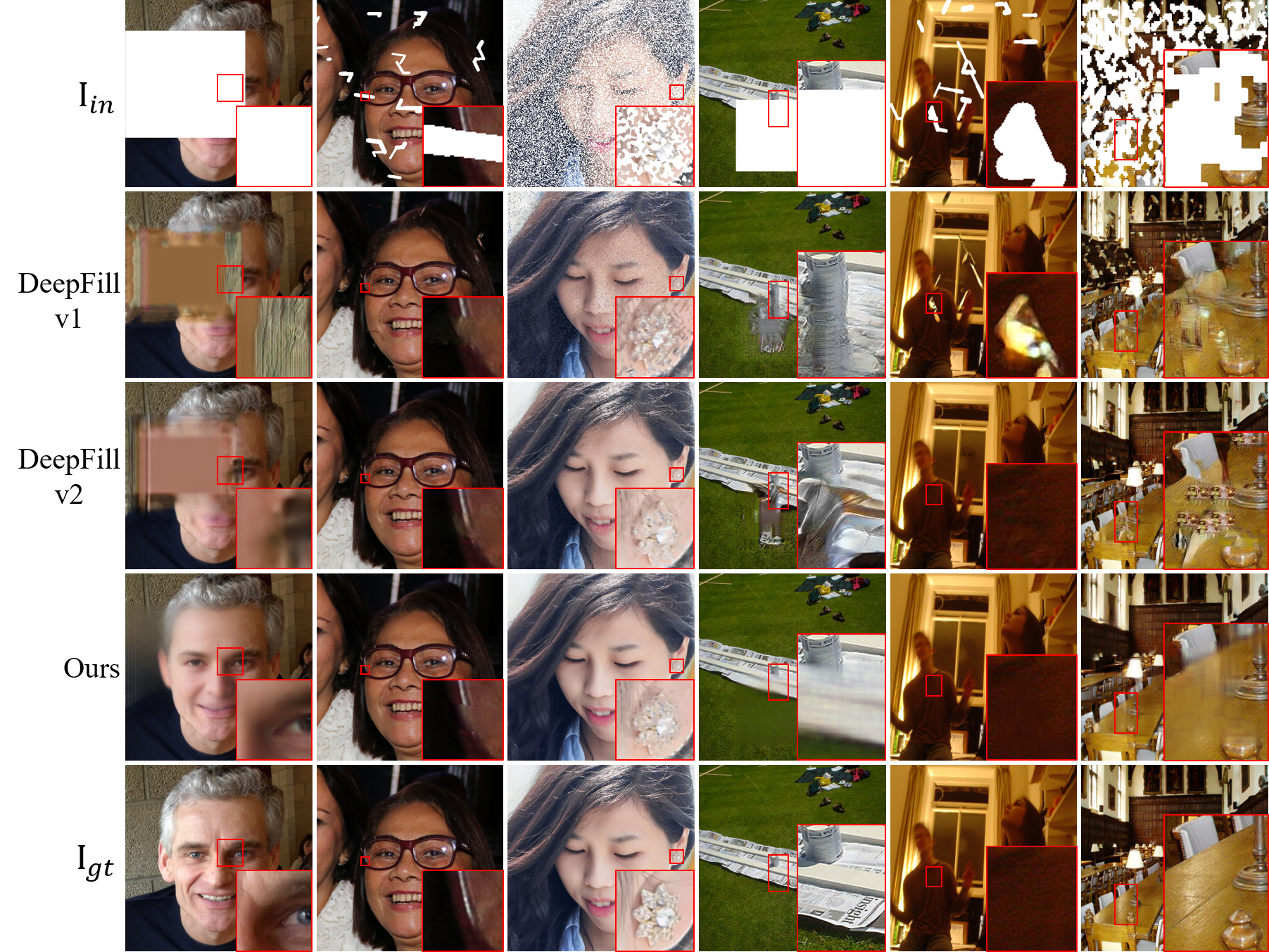}
\vspace{-1\baselineskip}
\caption{\textbf{Comparisons of test results on FFHQ and Oxford Buildings datasets.} Each column shows an example of the test results. From top to bottom: the first row displays various masked input images from both datasets. The second to the fourth rows show the completed images by DeepFillv1, v2 and our DeepGIN. The reference ground truth images are also provided at the last row. Zoom in for a better view}
\label{fig:comparisons_ffhq_oxford}
\vspace{-1.5\baselineskip}
\end{figure}

\vspace{-1.25\baselineskip}
\subsubsection{Quantitative Comparisons.} Table~\ref{table:comparisons} shows the comparisons with DeepFillv1 and v2 on the two datasets with two sets of masked images. It is clear that our model outperforms DeepFillv1 and v2 in all the experiments on the two datasets in terms of the pixel-wise reconstruction accuracy. Our model achieves better PSNR compared with DeepFillv1 and v2 in the range of 1.84$\sim$5.1 dB and offers better SSIM and L1 error. For FID and LPIPS, we attain better performance on the testing images with the original sizes. For the testing images with size of 256$\times$256 and masked by random rectangular masks, we are also comparable to the other two approaches.

\vspace{-1.25\baselineskip}
\subsubsection{Qualitative Comparisons.} 
\begin{figure}[t]
\centering
\includegraphics[width=0.85\textwidth]{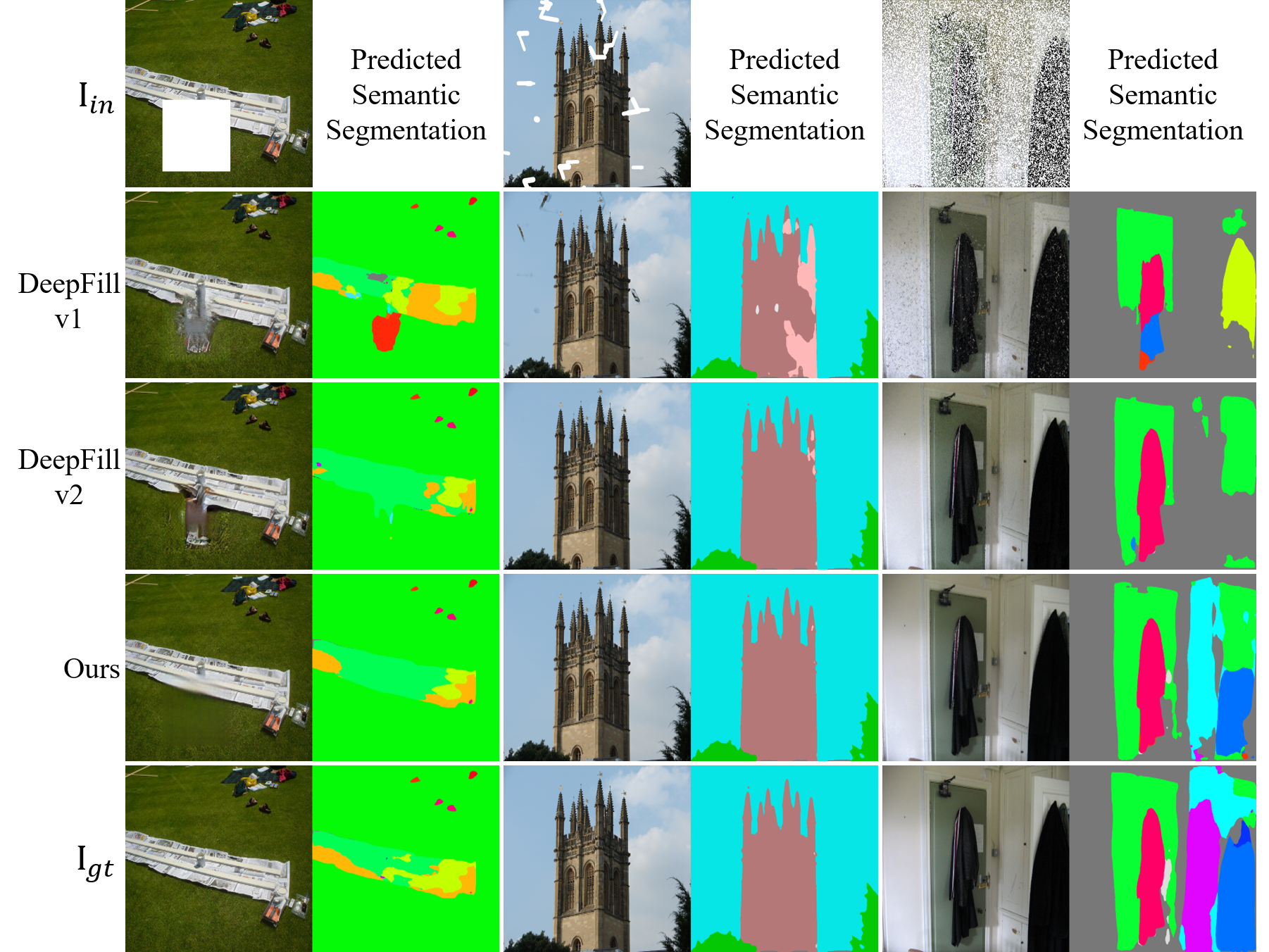}
\vspace{-1\baselineskip}
\caption{\textbf{Visualizations of predicted semantic segmentation test results on Oxford Buildings dataset.} $2^{\mathrm{nd}}$ to $4^{\mathrm{th}}$ rows show the completed images by different methods and the corresponding predicted semantic segmentation obtained using the trained network \cite{ADE20k2}. The ground truth images are also attached to the last row for reference. Please zoom in for a better view}
\label{fig:visualization_seg}
\vspace{-1.5\baselineskip}
\end{figure}

Fig.~\ref{fig:comparisons_ffhq_oxford} displays the test results on both FFHQ and Oxford datasets. It can be seen that DeepFillvl and v2 fail to achieve satisfactory visual quality on the large rectangular masks as shown in the first and fourth columns in Fig.~\ref{fig:comparisons_ffhq_oxford}. For the other two types of masked images, our model also provides the completed images with better color and content coherency. Note that our model tends to produce blurry images and the reason is that our model was trained to be more PSNR-oriented than the DeepFillv1 and v2. We seek a balance between the pixel-wise accuracy and the visual quality to avoid some strange generated patterns like the completed image by DeepFillv2 of the last example in Fig.~\ref{fig:comparisons_ffhq_oxford}. To show that our model offers better pixel-wise accuracy, we provide the predicted semantic segmentation test results in Fig.~\ref{fig:visualization_seg}. It is obvious that our results are semantically closer to $\textbf{I}_{gt}$ than that of the other two methods, see for example, the intersection of the newspaper and the lawn in Fig.~\ref{fig:visualization_seg}.

\vspace{-0.75\baselineskip}
\section{Conclusions}

\vspace{-0.75\baselineskip}
We have presented a deep generative inpainting network, called DeepGIN. Unlike the existing works, we propose a Spatial Pyramid Dilation (SPD) ResNet block to include more receptive fields for utilizing information given by distant spatial locations. This is important to inpainting especially when the masked regions are too large to be filled. We also enhance the significance of self-similarity consideration, hence we employ Multi-Scale Self-Attention (MSSA) strategy to enhance our performance. Furthermore, Back Projection (BP) is strategically used to improve the alignment of the generated and valid pixels. We have achieved performance better than the state-of-the-art image inpainting. This research work participated in the AIM 2020 Extreme Image Inpainting Challenge, which requires the right balance of pixel-wise reconstruction accuracy and visual quality. We believe that our DeepGIN is able to achieve the right balance and we encourage scholars in the field to give more attention in this direction.

\clearpage
%
%
\bibliographystyle{splncs04}
\bibliography{egbib}

\begin{thebibliography}{10}
\providecommand{\url}[1]{\texttt{#1}}
\providecommand{\urlprefix}{URL }
\providecommand{\doi}[1]{https://doi.org/#1}

\bibitem{AIM2020}
AIM2020: Aim 2020 extreme image inpainting challenge: Methods and results. In:
  ECCV Workshops (2020)

\bibitem{ToDayGAN}
Anoosheh, A., Sattler, T., Timofte, R., Pollefeys, M., Gool, L.V.: Night-to-day
  image translation for retrieval-based localization. 2019 International
  Conference on Robotics and Automation (ICRA)  (May 2019).
  \doi{10.1109/icra.2019.8794387}

\bibitem{PatchMatch}
Barnes, C., Shechtman, E., Finkelstein, A., Goldman, D.B.: Patchmatch: A
  randomized correspondence algorithm for structural image editing. ACM Trans.
  Graph.  \textbf{28}(3), ~24 (2009)

\bibitem{StyleTransfer}
Gatys, L.A., Ecker, A.S., Bethge, M.: A neural algorithm of artistic style.
  arXiv preprint arXiv:1508.06576  (2015)

\bibitem{GANs}
Goodfellow, I., Pouget-Abadie, J., Mirza, M., Xu, B., Warde-Farley, D., Ozair,
  S., Courville, A., Bengio, Y.: Generative adversarial nets. In: Advances in
  neural information processing systems. pp. 2672--2680 (2014)

\bibitem{AIM19-SR}
Gu, S., Danelljan, M., Timofte, R., Haris, M., Akita, K., Shakhnarovic, G.,
  Ukita, N., Michelini, P.N., Chen, W., Liu, H., et~al.: Aim 2019 challenge on
  image extreme super-resolution: Methods and results. In: 2019 IEEE/CVF
  International Conference on Computer Vision Workshop (ICCVW). pp. 3556--3564.
  IEEE (2019)

\bibitem{DBPN}
Haris, M., Shakhnarovich, G., Ukita, N.: Deep back-projection networks for
  super-resolution. In: Proceedings of the IEEE conference on computer vision
  and pattern recognition. pp. 1664--1673 (2018)

\bibitem{ResNet}
He, K., Zhang, X., Ren, S., Sun, J.: Deep residual learning for image
  recognition. In: Proceedings of the IEEE conference on computer vision and
  pattern recognition. pp. 770--778 (2016)

\bibitem{FID}
Heusel, M., Ramsauer, H., Unterthiner, T., Nessler, B., Hochreiter, S.: Gans
  trained by a two time-scale update rule converge to a local nash equilibrium.
  In: Advances in neural information processing systems. pp. 6626--6637 (2017)

\bibitem{GLCIC}
Iizuka, S., Simo-Serra, E., Ishikawa, H.: Globally and locally consistent image
  completion. ACM Trans. Graph.  \textbf{36}(4) (Jul 2017).
  \doi{10.1145/3072959.3073659}

\bibitem{pix2pix}
Isola, P., Zhu, J.Y., Zhou, T., Efros, A.A.: Image-to-image translation with
  conditional adversarial networks. In: Proceedings of the IEEE conference on
  computer vision and pattern recognition. pp. 1125--1134 (2017)

\bibitem{Johnson}
Johnson, J., Alahi, A., Fei-Fei, L.: Perceptual losses for real-time style
  transfer and super-resolution. In: European conference on computer vision.
  pp. 694--711. Springer (2016)

\bibitem{CelebA-HQ}
Karras, T., Aila, T., Laine, S., Lehtinen, J.: Progressive growing of gans for
  improved quality, stability, and variation. arXiv preprint arXiv:1710.10196
  (2017)

\bibitem{FFHQ}
Karras, T., Laine, S., Aila, T.: A style-based generator architecture for
  generative adversarial networks. In: Proceedings of the IEEE conference on
  computer vision and pattern recognition. pp. 4401--4410 (2019)

\bibitem{Adam}
Kingma, D.P., Ba, J.: Adam: A method for stochastic optimization. arXiv
  preprint arXiv:1412.6980  (2014)

\bibitem{ITSC}
{Li}, C., {Siu}, W., {Lun}, D.P.K.: Vision-based place recognition using
  convnet features and temporal correlation between consecutive frames. In:
  2019 IEEE Intelligent Transportation Systems Conference (ITSC). pp.
  3062--3067 (2019)

\bibitem{PartialConv}
Liu, G., Reda, F.A., Shih, K.J., Wang, T.C., Tao, A., Catanzaro, B.: Image
  inpainting for irregular holes using partial convolutions. In: Proceedings of
  the European Conference on Computer Vision (ECCV). pp. 85--100 (2018)

\bibitem{APBN}
Liu, Z.S., Wang, L.W., Li, C.T., Siu, W.C., Chan, Y.L.: Image super-resolution
  via attention based back projection networks. In: 2019 IEEE/CVF International
  Conference on Computer Vision Workshop (ICCVW). pp. 3517--3525. IEEE (2019)

\bibitem{NTIRE20-SR}
Lugmayr, A., Danelljan, M., Timofte, R.: Ntire 2020 challenge on real-world
  image super-resolution: Methods and results. In: Proceedings of the IEEE/CVF
  Conference on Computer Vision and Pattern Recognition Workshops. pp. 494--495
  (2020)

\bibitem{SN-GANs}
Miyato, T., Kataoka, T., Koyama, M., Yoshida, Y.: Spectral normalization for
  generative adversarial networks. arXiv preprint arXiv:1802.05957  (2018)

\bibitem{EdgeConnect}
Nazeri, K., Ng, E., Joseph, T., Qureshi, F.Z., Ebrahimi, M.: Edgeconnect:
  Generative image inpainting with adversarial edge learning. arXiv preprint
  arXiv:1901.00212  (2019)

\bibitem{Pytorch}
Paszke, A., Gross, S., Massa, F., Lerer, A., Bradbury, J., Chanan, G., Killeen,
  T., Lin, Z., Gimelshein, N., Antiga, L., et~al.: Pytorch: An imperative
  style, high-performance deep learning library. In: Advances in neural
  information processing systems. pp. 8026--8037 (2019)

\bibitem{CE}
Pathak, D., Krahenbuhl, P., Donahue, J., Darrell, T., Efros, A.A.: Context
  encoders: Feature learning by inpainting. In: Proceedings of the IEEE
  conference on computer vision and pattern recognition. pp. 2536--2544 (2016)

\bibitem{Oxford}
Philbin, J., Chum, O., Isard, M., Sivic, J., Zisserman, A.: Object retrieval
  with large vocabularies and fast spatial matching. In: 2007 IEEE conference
  on computer vision and pattern recognition. pp.~1--8. IEEE (2007)

\bibitem{PixelShuffle}
Shi, W., Caballero, J., Husz{\'a}r, F., Totz, J., Aitken, A.P., Bishop, R.,
  Rueckert, D., Wang, Z.: Real-time single image and video super-resolution
  using an efficient sub-pixel convolutional neural network. In: Proceedings of
  the IEEE conference on computer vision and pattern recognition. pp.
  1874--1883 (2016)

\bibitem{VGG-19}
Simonyan, K., Zisserman, A.: Very deep convolutional networks for large-scale
  image recognition. arXiv preprint arXiv:1409.1556  (2014)

\bibitem{DLN}
Wang, L.W., Liu, Z.S., Siu, W.C., Lun, D.P.: Lightening network for low-light
  image enhancement. IEEE Transactions on Image Processing  (2020).
  \doi{10.1109/TIP.2020.3008396}

\bibitem{pix2pixHD}
Wang, T.C., Liu, M.Y., Zhu, J.Y., Tao, A., Kautz, J., Catanzaro, B.:
  High-resolution image synthesis and semantic manipulation with conditional
  gans. In: Proceedings of the IEEE conference on computer vision and pattern
  recognition. pp. 8798--8807 (2018)

\bibitem{NonLocalBlk}
Wang, X., Girshick, R., Gupta, A., He, K.: Non-local neural networks. In:
  Proceedings of the IEEE conference on computer vision and pattern
  recognition. pp. 7794--7803 (2018)

\bibitem{ESRGAN}
Wang, X., Yu, K., Wu, S., Gu, J., Liu, Y., Dong, C., Qiao, Y., Change~Loy, C.:
  Esrgan: Enhanced super-resolution generative adversarial networks. In:
  Proceedings of the European Conference on Computer Vision (ECCV). pp.~0--0
  (2018)

\bibitem{SSIM}
Wang, Z., Bovik, A.C., Sheikh, H.R., Simoncelli, E.P.: Image quality
  assessment: from error visibility to structural similarity. IEEE transactions
  on image processing  \textbf{13}(4),  600--612 (2004)

\bibitem{MSNPS}
Yang, C., Lu, X., Lin, Z., Shechtman, E., Wang, O., Li, H.: High-resolution
  image inpainting using multi-scale neural patch synthesis. In: Proceedings of
  the IEEE Conference on Computer Vision and Pattern Recognition. pp.
  6721--6729 (2017)

\bibitem{CA}
Yu, J., Lin, Z., Yang, J., Shen, X., Lu, X., Huang, T.S.: Generative image
  inpainting with contextual attention. In: Proceedings of the IEEE conference
  on computer vision and pattern recognition. pp. 5505--5514 (2018)

\bibitem{Deepfillv2}
Yu, J., Lin, Z., Yang, J., Shen, X., Lu, X., Huang, T.S.: Free-form image
  inpainting with gated convolution. In: Proceedings of the IEEE International
  Conference on Computer Vision. pp. 4471--4480 (2019)

\bibitem{LPIPS}
Zhang, R., Isola, P., Efros, A.A., Shechtman, E., Wang, O.: The unreasonable
  effectiveness of deep features as a perceptual metric. In: Proceedings of the
  IEEE conference on computer vision and pattern recognition. pp. 586--595
  (2018)

\bibitem{ADE20K1}
Zhou, B., Zhao, H., Puig, X., Fidler, S., Barriuso, A., Torralba, A.: Scene
  parsing through ade20k dataset. In: Proceedings of the IEEE Conference on
  Computer Vision and Pattern Recognition (2017)

\bibitem{ADE20k2}
Zhou, B., Zhao, H., Puig, X., Xiao, T., Fidler, S., Barriuso, A., Torralba, A.:
  Semantic understanding of scenes through the ade20k dataset. International
  Journal on Computer Vision  (2018)

\end{thebibliography}

\end{document}